\documentclass[runningheads,a4paper]{llncs}

\usepackage{amssymb}
\usepackage{amsmath}
\usepackage{graphicx}

\usepackage{url}
\urldef{\mailsa}\path|Wiem.Maalel@gmail.com|
\urldef{\mailsb}\path|kzhoumath@163.com|
\urldef{\mailsc}\path|Arnaud.Martin@univ-rennes1.fr|
\urldef{\mailsd}\path|Zied.Elouedi@gmx.fr|
\newcommand{\keywords}[1]{\par\addvspace\baselineskip
\noindent\keywordname\enspace\ignorespaces#1}

\newcommand{\betP}{\mathrm{BetP}}

\newcommand{\ocap}{\displaystyle{\small{\textcircled{{\scriptsize $\cap$}}}}}
\newcommand{\eR}{\mbox{I\hspace{-.15em}R}}
\newcommand{\argmax}{\operatornamewithlimits{arg\,max}}

\begin{document}

\mainmatter  % start of an individual contribution

\title{Belief Hierarchical Clustering}

%\titlerunning{Lecture Notes in Computer Science: Authors' Instructions}

\author{Wiem Maalel
\and Kuang Zhou\and Arnaud Martin\and Zied Elouedi}
\authorrunning{Belief Hierarchical Clustering}

\institute{LARODEC, ISG.41 rue de la Libert\'e, 2000 Le Bardo, Tunisia\\
IRISA, Universit\'e de Rennes 1. IUT de Lannion. Rue Edouard Branly BP 30219, 22302 Lannion cedex. France \\
\mailsa\\
\mailsb\\
\mailsc\\
\mailsd\\
}

\toctitle{Belief Hierarchical Clustering}
\tocauthor{Wiem Maalel, Kuang Zhou, Arnaud Martin, Zied Elouedi}
\maketitle

\begin{abstract}
In the data mining field many clustering methods have been proposed, yet standard versions do not take into account uncertain databases. This paper deals with a new approach to cluster uncertain data by using a hierarchical clustering defined within the belief function framework. The main objective of the belief hierarchical clustering is to allow an object to belong to one or several clusters. To each belonging, a degree of belief is associated, and clusters are combined based on the pignistic properties. Experiments with real uncertain data show that our proposed method can be considered as a propitious tool.
\end{abstract}

\section{Introduction}
\label{INT}
Due to the increase of imperfect data, the process of decision making is becoming harder. In order to face this, the data analysis is being applied in various fields.

Clustering is mostly used in data mining and aims at grouping a set of similar objects into clusters. In this context, many clustering algorithms exist and are categorized into two main families:\\
The first family involves the partitioning methods based on density such as $k$-means algorithm \cite{MacQueen67a} that is widely used thanks to its convergence speed. It partitions the data into $k$ clusters represented by their centers. The second family includes the hierarchical clustering methods such as the top-down and
%Besides, it is characterized by its sensitivity to noisy data and outliers.
the Hierarchical Ascendant Clustering (HAC) \cite{Hastiea}. This latter consists on constructing clusters recursively by partitioning the objects in a bottom-up way. This process leads to good result visualizations. Nevertheless, it has a non-linear complexity.
%We found also the mixed clustering of the K-means method, applied at the first and the final step, and the hierarchical clustering, applied at the second step. This method deals with data tables containing a thousands of individuals.

All these standard methods deal with certain and precise data. Thus, in order to facilitate the decision making, it would be more appropriate to handle uncertain data. Here, we need a soft clustering process that will take into account the possibility that objects belong to more than one cluster.
%Furthermore, when information about the data uncertainty is available, it can improve the quality of the results.

In such a case, several methods have been established. Among them, the Fuzzy C-Means \cite{Bezdek03a} which consists in assigning a membership to each data point corresponding to the cluster center, and the weights minimizing the total weighted mean-square error. This method constantly converges. Patently, Evidential $c$-Means (ECM) \cite{Denoeux04a}, \cite{Masson04a} is deemed to be a very fateful method. It enhances the FCM and generates a credal partition from attribute data. This method deals with the clustering of object data. Accordingly, the belief $k$-Modes method \cite{Sarra06a} is a popular method, which builds K groups characterized by uncertain attribute values and provides a classification of new instances. Schubert has also found a clustering algorithm \cite{Schubert03} which uses the mass on the empty set to build a classifier.
%so that, an object may belong to one or several clusters.

%but it has a sensitivity to initialization and a weakness face to noisy data.
%and is characterized by its robustness to inaccuracies and outliers. Though, it does not take into account a priori knowledge.
Our objective in this paper is to develop a belief hierarchical clustering method, in order to ensure the membership of objects in several clusters, and to handle the uncertainty in data under the belief function framework.

This remainder is organized as follows: in the next section we review the ascendant hierarchical clustering, its concepts and its characteristics. In section 3, we recall some of the basic concepts of belief function theory. Our method is described in section 4 and we evaluate its performance on a real data set in section 5. Finally, Section 6 is a conclusion for the whole paper.
\section{Ascendant hierarchical clustering}
\label{CAH}
This method  consists on agglomerating the close clusters in order to have finally one cluster containing all the objects $x_j$ (where $j=1,..,N)$.\\
Let's consider $\mathcal{P}^{K}=\{C_1,...,C_K\}$ the set of clusters. If $K=N$, $C_1=x_1,...,C_N=x_N$. Thereafter, throughout all the steps of clustering, we will move from a partition $\mathcal{P}^{K}$ to a partition $\mathcal{P}^{K-1}$.
The result generated is described by a hierarchical clustering tree (dendrogram), where the nodes represent the successive fusions and the height of the nodes represents the value of the distance between two objects which gives a concrete meaning to the level of nodes conscripted as "indexed hierarchy". This latter is usually indexed by the values of the distances (or dissimilarity) for each aggregation step. The indexed hierarchy can be seen as a set with an ultrametric distance $d$ which satisfies these properties:
{\it i)} $x=y \Longleftrightarrow d(x,y)=0$.\\
{\it ii)} $d(x,y)=d(y,x)$.\\
{\it iii)} $d(x,y)\leq d(x,z)+d(y,z), \forall x, y, z \in \eR$.\\

The algorithm is as follows:
\begin{itemize}
 \item Initialisation: the initial clusters are the N-singletons. We compute their dissimilarity matrix.
 \item Iterate these two steps until the aggregation turns into a single cluster:
 \begin{itemize}
  \item Combine the two most similar (closest) elements (clusters) from the selected groups according to some distance rules.
  \item Update the  matrix distance by replacing the two grouped elements by the new one and calculate its distance from each of the other classes.
 \end{itemize}

\end{itemize}

Once all these steps completed, we do not recover a partition of $K$ clusters, but a partition of $K-1$ clusters. Hence, we had to point out the aggregation criterion (distance rules) between two points and between two clusters.
%AM A simplifier et bien expliquer
%\textbf{Distance between two objects:}
We can use the Euclidian distance between $N$ objects $x$ defined in a space $\eR$. Different distances can be considered between two clusters: we can consider the minimum as follows:
\begin{equation}
\textbf{d}(C^i_j,C^i_{j'})= \min_{x_k\in C^i_j,x_{k'}\in C^i_{j'}} d(x_k,x_{k'})
\end{equation}
with $j,j'=1,...,i$. The maximum can also be considered, however, the minimum and maximum distances create compact clusters but sensitive to "outliers". The average can also be used, but the most used method is Ward's method, using Huygens formula to compute this:
\begin{equation}
\Delta I_{inter(C^i_j,C^i_{j'})}=\frac{m_{C_j} m_{C_{j'}}}{m_{C_j}+m_{C_{j'}}} d^2({\bf C^i_j,C^i_{j'}})
\end{equation}
where $m_{C_j}$ and $m_{C_{j'}}$ are numbers of elements of $C_j$ and $C_{j'}$ respectively and ${\bf C^i_j,C^i_{j'}}$ the centers. Then, we had to find the couple of clusters minimizing the distance:
\begin{equation}
(C^k_l,C^k_{l'})=d(C^i_l,C^i_{l'})=\min_{C^i_j,C^i_{l'}\in \mathcal{C}^i} \textbf{d}(C^i_j,C^i_{j'})
\label{moneq}
\end{equation}

% We can use several distance formulae to compute $\textbf{d} (C^i_j,C^i_{j'})$:\\
% \textbf{Single-link =} similarity of the two most similar members, in other words, we choose the minimum of the distance between two objects : creates elongated clusters and a single pair of members close enough: \\
% \begin{equation}
% D_{C^i_j,C^i_{j'}}=\min_{x_i\in C^i_j,i'\in C^i_{j'}} (d_{C^i_j,C^i_{j'}})
% \end{equation}
% \textbf{ Complete-link =} similarity of the two most dissimilar members, the maximum distance is used: creates compact clusters and sensitive to "outliers":\\
% \begin{equation}
% D_{C^i_j,C^i_{j'}}=\max_{x_i\in C^i_j,i'\in C^i_{j'}} (d_{C^i_j,C^i_{j'}})
% \end{equation}
% \textbf{Group-average =} average similarity of all pairs of members.\\
% \begin{equation}
% D_{avg}(C^i_j,C^i_{j'})=\frac{1}{\mid C^i_j\mid \times \mid C^i_{j'}\mid} \sum_{x\in A,y \in B}\limits d(x,y)
% \end{equation}
% \textbf{Ward's method}: we use Huygens formula to compute this:\\
% \begin{equation}
% \Delta I_{inter(C^i_j,C^i_{j'})}=\frac{m_{C_j} m_{C_{j'}}}{m_{C_j}+m_{C_{j'}}} d^2(C^i_j,C^i_{j'})
% \end{equation}
% where $m_{C_j}$ and $m_{C_{j'}}$ are masses of classes $C_j$ and $C_{j'}$ respectively.\\

% Presenter dendrogram
\section{Basis on the theory of belief functions}
\label{BF}
In this Section, we briefly review the main concepts that will be used in our method that underlies the theory of
belief functions \cite{Shafera} as interpreted in the Transferable Belief Model (TBM) \cite{Smets94a}. Let's suppose that the frame of discernment is $\Omega= \left\{\omega_1, \omega_2,...,\omega_3\right\}$. $\Omega$ is a finite set that reflects a state of partial knowledge that can be represented by a basis belief assignment defined as:
\begin {equation}
\begin {array}{ccc}
m:2^\Omega\rightarrow[0,1]\\
\sum_{A\subseteq \Omega}\limits m(A)=1
\end {array}
\end {equation}
The value $m(A)$ is named a basic belief mass (bbm) of $A$. The subset $A \in 2^\Omega$ is called focal element if $m(A)> 0$.
%The belief function bel(A) expresses the degree of belief in the hypothesis $"'w\in A"$, given the belief masses assigned to all cases involving A, whereas, the plausibility function expresses the maximum degree of belief that can potentially be attributed to the hypothesis $"w\in A"$.
% \begin {equation}
% \begin {array}{ccc}
% bel:2^\Omega \mapsto[0,1]\\
%     A\rightarrow\sum_{B\subseteq A,B\neq\emptyset}\limits m(B)
% \end{array}
% \end {equation}
% \begin {equation}
% \begin {array}{ccc}
% pl:2^\Omega \mapsto[0,1]\\
%     A\rightarrow\sum_{B\cap A\neq\emptyset}\limits m(B)
% \end {array}
% \end {equation}
One of the important rules in the belief theory is the conjunctive rule which consists on combining two basic belief assignments $m_1$ and $m_2$ induced from two distinct and reliable information sources defined as:
\begin{equation}
        \label{SmetsRule}
            m_{1} \ocap m_2(C)=\sum _{A\cap B=C} m_1(A)\cdot m_2(B), \quad \forall C \subseteq \Omega
    \end{equation}
The Dempster rule is the normalized conjunctive rule:
\begin{equation}
        \label{DempsterRule}
            m_{1} \oplus m_2(C)=\frac{ m_{1} \ocap m_2(C)}{1- m_{1} \ocap m_2(\emptyset)}, \quad \forall C \subseteq \Omega
    \end{equation}

In order to ensure the decision making, beliefs are transformed into probability measures recorded $\betP$, and defined as follows \cite{Smets94a}:
\begin{equation}
\betP(A)=\sum_{B\subseteq\Omega}\frac{|A\cap B\mid}{\mid B\mid}\frac{m(B)}{(1-m(\emptyset))}, \forall A \in\Omega
\end{equation}

\section{Belief  hierarchical clustering}
\label{BHC}
In order to set down a way to develop a belief hierarchical clustering, we choose to work on different levels: on one hand, the object level, on the other hand, the cluster level. At the beginning, for $N$ objects we have, the frame of discernment is  \linebreak $\Omega=\left\{x_1,...,x_N\right\}$ and for each object belonging to one cluster, a degree of belief is assigned. Let $\mathcal{P}^N$ be the partition of $N$ objects. Hence, we define a mass function for each object $x_i$, inspired from the $k$-nearest neighbors \cite{Denoeux95a} method which is defined as follows:
\begin{equation}
\begin{array}{ccc}
m^{\Omega_i}_i(x_j)&=&\alpha e^{-\gamma d^2(x_i,x_j)}\\
m^{\Omega_i}_i(\Omega_i)&=&1-\sum m^{\Omega_i}_i(x_j)
\end{array}
\end{equation}
where $i\neq j$, $\alpha$ and $\gamma$ are two parameters we can optimize~\cite{Zouhal98a}, $d$ can be considered as the Euclidean distance, and the frame of discernment is given by $\Omega_i=\left\{x_1,...,x_N\right\}\backslash \left\{x_i\right\}$.

In order to move from the partition of $N$ objects to a partition of $N-1$ objects we have to find both nearest objects $(x_i,x_j)$ to form a cluster. Eventually, the partition of $N-1$ clusters will be given by $\mathcal{P}^{N-1}=\left\{(x_i,x_j),x_k\right\}$ where $k=1,...,N\backslash\left\{i,j\right\}$. The nearest objects are found considering the pignistic probability, defined on the frame $\Omega_i$, of each object $x_i$, where we proceed the comparison by pairs, by computing firstly the pignistic for each object, and then we continue the process using $\argmax$. The nearest objects are chosen using the maximum of the pignistic values between pairs of objects, and we will compute the product pair one by one.
\begin{equation}
(x_i,x_j)=\argmax_{x_i,x_j \in \mathcal{P}^{N}} \betP^{\Omega_i}_i(x_j)\ast \betP^{\Omega_j}_j(x_i)
\end{equation}

Then, this first couple of objects is a cluster. Now consider that we have a partition $\mathcal{P}^K$ of $K$ clusters $\left\{C_1, \ldots ,C_K\right\}$. In order to find the best partition $\mathcal{P}^{K-1}$ of $K-1$ clusters, we have to find the best couple of clusters to be merged. First, if we consider one of the classical distances $\bf{d}$ (single link, complete link, average, etc), presented in section~\ref{CAH}, between the clusters, we delineate a mass function, defined within the frame $\Omega_i$ for each cluster $C_i \in \mathcal{P}^K$ with $C_i \neq C_j$ by:
\begin{eqnarray}
m^{\Omega_i}_i(C_{j})&=&\alpha e^{-\gamma d^2(C_i,C_j)}\\
m^{\Omega_i}_i(\Omega_i)&=&1-\sum m^{\Omega_i}_i(C_j)
\end{eqnarray}
where $\Omega_i=\left\{C_1, \ldots ,C_K\right\}\setminus \{C_i\}$. Then, both clusters to merge are given by:
\begin{equation}
\label{PartDec}
(C_i,C_j)=\argmax_{C_i,C_j \in \mathcal{P}^{K}} \betP^{\Omega_i}(C_j)\ast \betP^{\Omega_j}(C_i)
\end{equation}
and the partition $\mathcal{P}^{K-1}$ is made from the new cluster $(C_i,C_j)$ and all the other clusters of $\mathcal{P}^{K}$. The point by doing so is to prove that if we maximize the degree of probability we will have the couple of clusters to combine. Of course, this approach will give exactly the same partitions than the classical ascendant hierarchical clustering, but the dendrogram can be built from $\betP$ %AM montrer le lien avec la discimilarit?et la hi?archie indic?
and the best partition ({\em i.e.} the number of clusters) can be preferred to find. The indexed hierarchy will be indexed by the sum of $\betP$ which will lead to more precise and specific results according to the dissimilarity between objects and therefore will facilitate our process.

Hereafter, we define another way to build the partition $\mathcal{P}^{K-1}$. For each initial object $x_i$ to classify, it exists a cluster of $\mathcal{P}^K$ such as $x_i \in C_k$. We consider the frame of discernment  $\Omega_i=\left\{C_1, \ldots ,C_K\right\}\setminus \{C_k\}$, $m$, which describes the degree that the two clusters could be merged, can be noted $m^\Omega$and we define the mass function:
\begin{eqnarray}
m^{\Omega_i}_i(C_{k_j})&=&\prod_{x_j\in C_{k_j}}\alpha e^{-\gamma d^2(x_i,x_j)}\\
m^{\Omega_i}_i(\Omega_i)&=& 1-\sum_{x_j\in C_{k_j}}\limits m^{\Omega_i}_i(C_{k_j})
\end{eqnarray}
In order to find a mass function for each cluster $C_i$ of $\mathcal{P}^{K}$, we combine all the mass functions given by all objects of $C_i$ by a combination rule such as the Dempster rule of combination given by equation~\eqref{DempsterRule}. Then, to merge both clusters we use the equation~\eqref{PartDec} as before. The sum of the pignisitic probabilities will be the index of the dendrogram, called $\betP$ index.

\section{Experimentations}
\begin{center} \begin{figure}[!thbt] \centering
\begin{tabular}{cc}
		 \includegraphics[width=0.45\linewidth, height=5cm]{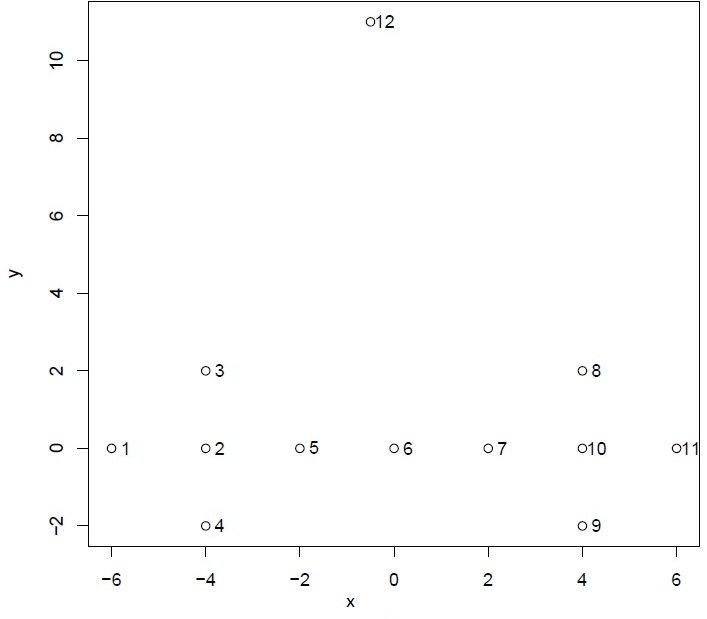} & \\
 {\centering\small a. Diamond data set} & \\
 \end{tabular}
 \begin{tabular}{cc}
		 \includegraphics[width=0.45\linewidth, height=5cm]{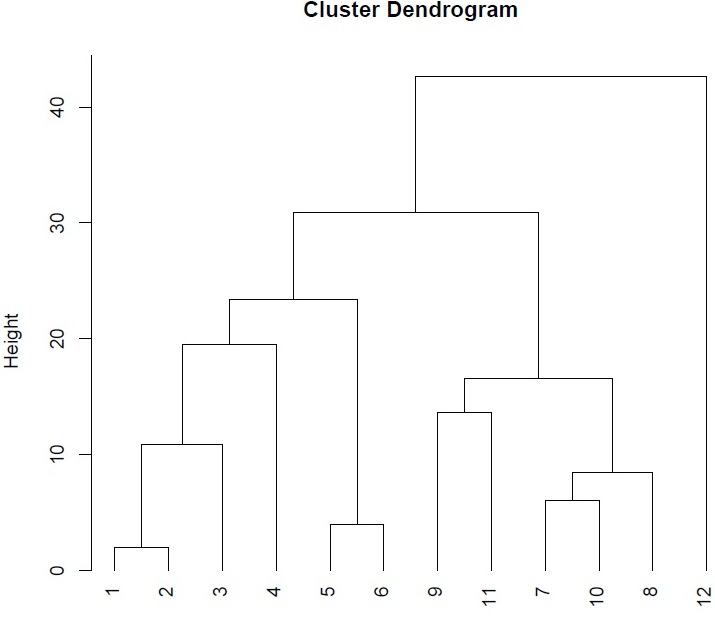} & 	 \includegraphics[width=0.45\linewidth, height=5cm]{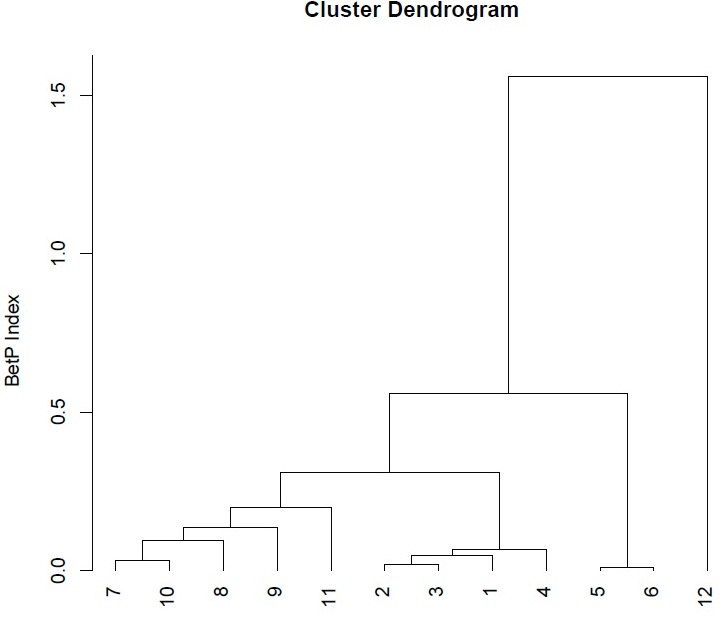} \\
{\centering\small b. Hierarchical clustering} & {\centering\small c. Belief hierarchical clustering} \\
		 \end{tabular}
				 \caption{Clustering results for Diamond data set.}
 \label{diamondRes} \end{figure} \end{center}

%\underline{Example 1\string:}\\
Experiments were first applied on diamond data set composed of twelve objects as describe in Figure~\ref{diamondRes}.a and analyzed in \cite{Masson04a}.
%In this data set, we are not obliged to apply the k-means method as a first step because the number of objects is not high, so we applied our hierarchical method directly.
The dendrograms for both classical and Belief Hierarchical Clustering (BHC) are represented by Figures~\ref{diamondRes}.b and~\ref{diamondRes}.c. The object 12 is well considered as an outlier with both approaches. With the belief hierarchical clustering, this object is clearly different, thanks to the pignistic probability. For HAC, the distance between object 12 and other objects is small, however, for BHC, there is a big gap between object 12 and others. This points out that our method is better for detecting outliers. If the objects 5 and 6 are associated to 1, 2, 3 and 4 with the classical hierarchical clustering, with BHC these points are more identified as different. This synthetic data set is special because of the equidistance of the points and there is no uncertainty.
% \begin{center} \begin{figure}[!thbt] \centering
% 		 \includegraphics[width=0.45\linewidth, height=5cm]{diamond.eps}
% 						 \caption{Diamond data set.}
%  \label{diamond} \end{figure} \end{center}
\newpage
	
 \begin{center} \begin{figure}[!thbt] \centering
\begin{tabular}{cc}
		 \includegraphics[width=0.45\linewidth, height=4cm]{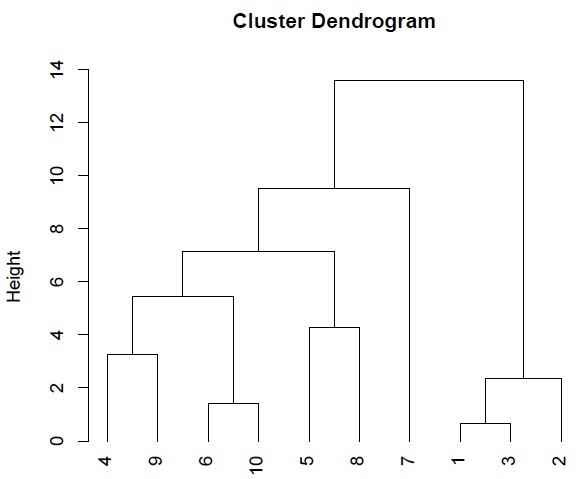} & 	 \includegraphics[width=0.45\linewidth, height=4cm]{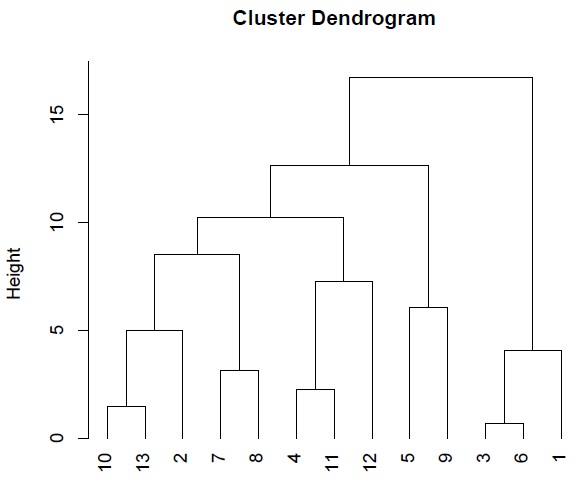} \\
{\centering\small a. $K_{init}=10$ for HAC} & {\centering\small b. $K_{init}=13$ for HAC} \\
		 \includegraphics[width=0.45\linewidth, height=4cm]{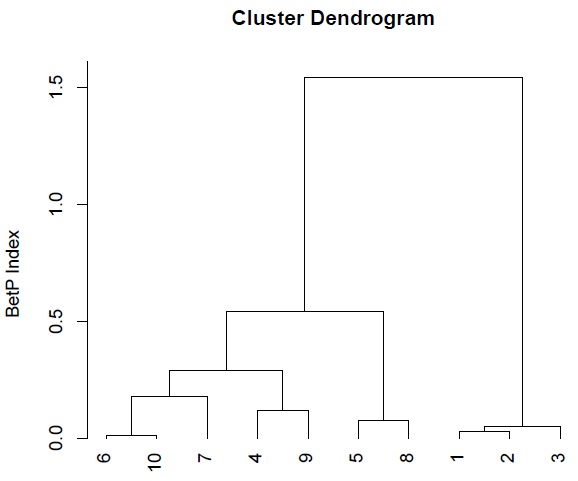} &  \includegraphics[width=0.45\linewidth, height=4cm]{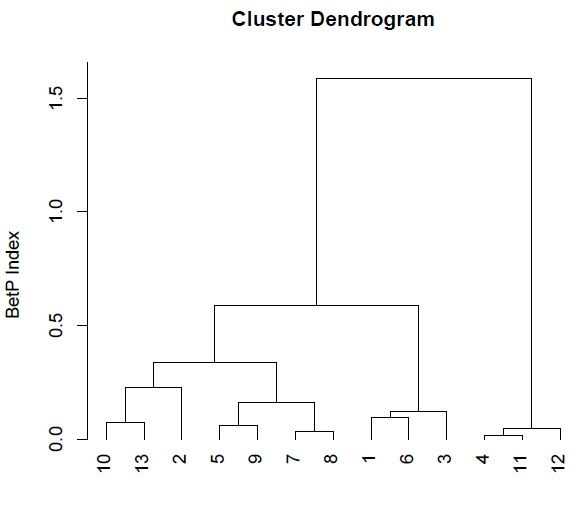} \\
{\centering\small c. $K_{init}=10$ for BHC} & {\centering\small d. $K_{init}=13$ for BHC} \\
	 \end{tabular}
		 \caption{Clustering results on IRIS data set for both hierarchical (HAC) (Fig. a and b) and belief hierarchical (BHC) (Fig. c and d) clustering ($K_{init}$ is the cluster number by $k$-means first).}
		 \label{1013} \end{figure} \end{center}

%\underline{Example 2\string:}\\
We continue our experiments with a well-known data set, Iris data set, which is composed of flowers from four types of species of Iris described by sepal length, sepal width, petal length, and petal width. %: Iris setosa, Iris versicolor, and Iris virginica. 50 observations for sepal length, sepal width, petal length, and petal width, exist from each species. In order to distinguish the species from each other, Fisher developed a linear discriminant model, based on the combination of the four features.
The data set contains three clusters known to have a significant overlap.

\iffalse
We increase then the  the initial cluster number by $k$-means from 10 to 13 and we notice that the best situation is also when the pignistic is equal to 0.3 as shown in the figure 2 because it describes the fact that the data set is composed of 3 clusters.
\fi
%For our method, we partition the Iris data set into three clusters containing respectively 50 objects.
In order to reduce the complexity and present distinctly the dendrogram, we first used the $k$-means method to get initial few clusters for our algorithm. It is not necessary to apply this method if the number of objects is not high.

%We will try to improve our approach to save space and time so we can apply BHC directly.
%, so that, the conjunctive combination of 50 or even 150 bbas in $2^{10}$ can be done without any problem of time consuming, and then, we use our method to get the dendogram.
Several experiments have been used with several number of clusters. We present in Figure~\ref{1013} the obtained dendrograms for 10 and 13 clusters. %We remark that at the beginning of the experiment the pignistic probability is too small and increases when the initial number of clusters increases, indicating the  clusters to be merged are more homogeneous.
We notice different combinations between the nearest clusters for both classical and belief hierarchical clustering. The best situation for BHC is obtained with the pignistic equal to $0.5$ because it indicates that the data set is composed of three significant clusters which reflects the real situation. For the classical hierarchical clustering the results are not so obvious. Indeed, for HAC, it is difficult to decide for the optimum cluster number because of the use of the euclidean distance and as seen in Figure 2.c it indistinguishable in terms of the y-value. However, for BHC, it is more easy to do this due to the use of the pignistic probability.   %it is small

%For 17 and 18, the dendrograms are almost the same, where i we notice that the best situation expresses the fact that the data set contains four clusters (see Figure 2). %The problem here is that it cannot work for more than 19 clusters not because of the method, however because it is due to the memory of the computer (we test using a 32-bit computer with 3GB memory).\\

In order to evaluate the performance of our method, we use some of the most popular measures: precision, recall and Rand Index (RI). The results for both BHC and HAC are summarized in Table 1. The first three columns are for BHC, while the others are for HAC. In fact, we suppose that $F_c$ represents the final number of clusters and we start with $F_c=2$ until $F_c=6$. We fixed the value of $k_{init}$ at 13. We note that for $F_c=2$ the precision is low while the recall is of high value, and that when we have a high cluster number ($F_c=5$ or 6), the precision will be high but the recall will be relatively low. Thus, we note that for the same number of final clusters ({\em e.g.} $F_c=4$), our method is better in terms of precision, recall and RI.
%Thus, in our method, $F_c=3$ clusters is the best case because both precision and recall are high along RI, and because it reflects the real situation for iris data set. However, $F_c=4$ cannot be considered as the best case because its RI is lower then the one of the previous case.
\newpage
\begin{table}[ht] \centering \caption{Evaluation results} \begin{tabular}{rrrrrrrr} \hline
      &  \multicolumn{ 3}{c}{BHC} &            & \multicolumn{ 3}{c}{HAC} \\
 & Precision & Recall & RI &~&  Precision & Recall & RI \\
  \hline
  $F_c=2$ & 0.5951 & 1.0000 & 0.7763 &~~~& 0.5951 & 1.0000 & 0.7763 \\
  $F_c=3$ & 0.8011 & 0.8438 & 0.8797 &~~~& 0.6079 & 0.9282 & 0.7795 \\
  $F_c=4$ & 0.9506 & 0.8275 & 0.9291 &~~~& 0.8183 & 0.7230 & 0.8561 \\
  $F_c=5$ & 0.8523 & 0.6063 & 0.8360 &~~~& 0.8523 & 0.6063 & 0.8360 \\
  $F_c=6$ & 0.9433 & 0.5524 & 0.8419 &~~~& 0.8916 & 0.5818 & 0.8392 \\
  \hline \end{tabular} \label{table}
 \end{table}

\begin{center} \begin{figure}[!thbt] \centering
\begin{tabular}{cc}
		 \includegraphics[width=0.45\linewidth, height=3.8cm]{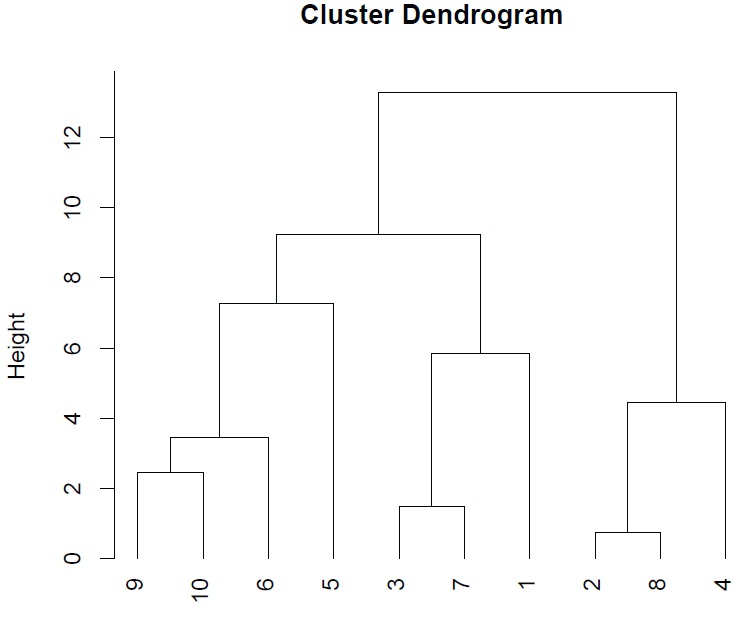} & 	 \includegraphics[width=0.45\linewidth, height=3.8cm]{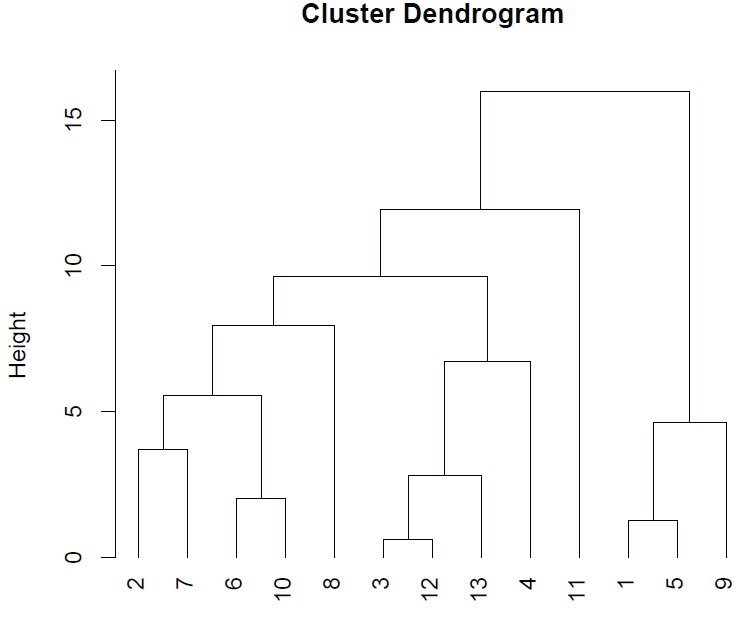} \\
{\centering\small a. $K_{init}=10$ for HAC} & {\centering\small b. $K_{init}=13$ for HAC} \\
		 \includegraphics[width=0.45\linewidth, height=3.8cm]{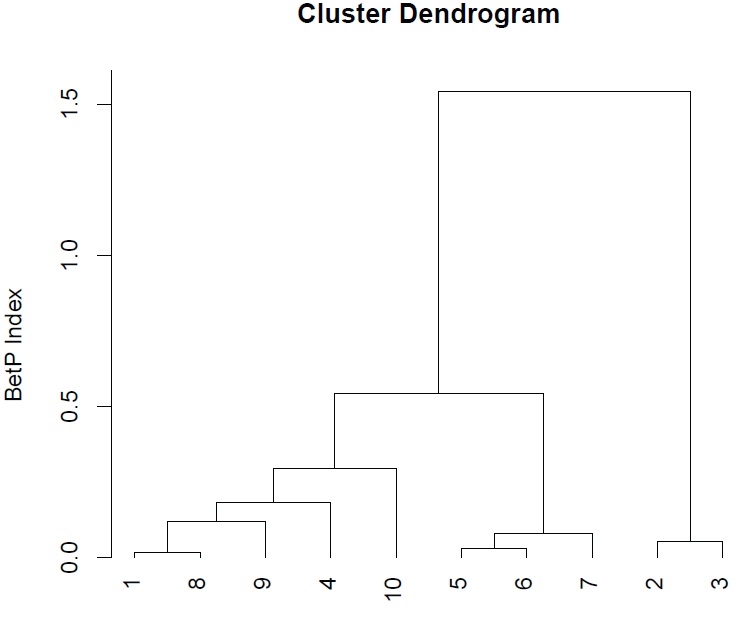} &  \includegraphics[width=0.45\linewidth, height=3.8cm]{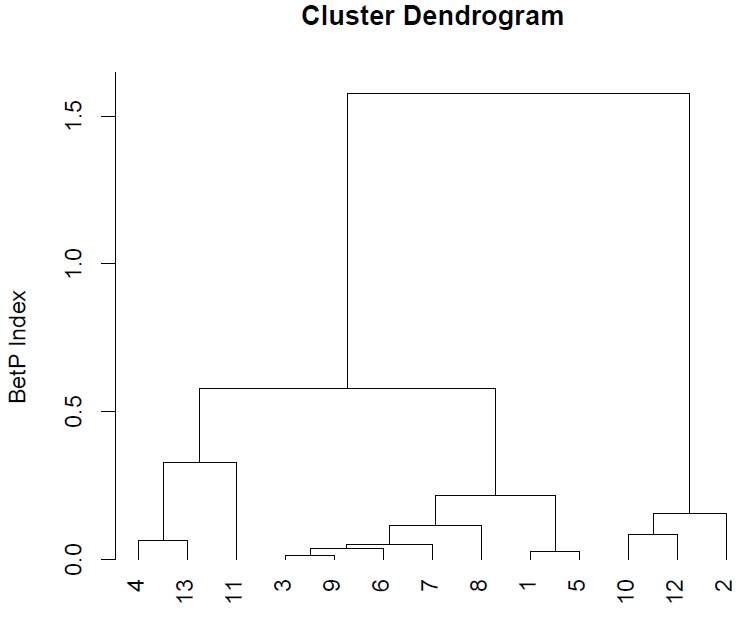} \\
{\centering\small c. $K_{init}=10$ for BHC} & {\centering\small d. $K_{init}=13$ for BHC} \\
	 \end{tabular}
		 \caption{Clustering results on Congressional Voting Records Data Set for both hierarchical and belief hierarchical clustering.}
		 \label{vote} \end{figure} \end{center}

Tests are also performed to a third data base, Congressional Voting Records Data Set. The results presented in Figure~\ref{vote} show that the pignistic probability value increased at each level, having thereby, a more homogeneous partition. We notice different combinations, between the nearest clusters, that are not the same within the two methods compared. For example, cluster 9 is associated to cluster 10 and then to 6 with HAC, but, with BHC it is associated to cluster 4 and then to 10. Although, throughout the BHC dendrograms shown in Figure~\ref{vote}.c and Figure~\ref{vote}.d, the best situation indicating the optimum number of clusters can be clearly obtained. This easy way is due to the use of the pignistic probability. %showing the best situation for this data set and that this latter can be represented using 3 clusters.
%\newpage
For this data set, we notice that for $F_c=2$ and 3, the precision is low while the recall is high. However, with the increasing of our cluster number, we notice that BHC provides a better results. In fact, for $F_c=3, 4, 5$ and 6 the precision and RI values relative to BHC are higher then the precision and RI values relative to HAC, which confirmed the efficiency of our approach which is better in terms of precision and RI.
		%\newpage
\begin{table}[ht]
\centering \caption{Evaluation results for Congressional Voting Records Data Set} \begin{tabular}{rrrrrrrr} \hline
    &  \multicolumn{ 3}{c}{BHC} &            & \multicolumn{ 3}{c}{HAC} \\
& Precision & Recall & RI &~&  Precision & Recall & RI \\
\hline
  $F_c=2$ & 0.3873 & 0.8177 & 0.5146 &~~~& 0.5951 & 1.0000 & 0.7763 \\
  $F_c=3$ & 0.7313 & 0.8190 & 0.8415 &~~~& 0.6288 & 0.8759 & 0.7892 \\
  $F_c=4$ & 0.8701 & 0.6833 & 0.8623 &~~~& 0.7887 & 0.7091 & 0.8419 \\
  $F_c=5$ & 0.8670 & 0.6103 & 0.8411 &~~~& 0.7551 & 0.6729 & 0.8207\\
  $F_c=6$ & 0.9731 & 0.6005 & 0.8632 &~~~& 0.8526 & 0.6014 & 0.8347 \\
   \hline
\end{tabular}
\end{table}
%\underline{Example 3\string:}\\
%In order to more evaluate our method, we apply on iris the mixed clustering which is a combination of the K-means method, applied at the first and final step, and the hierarchical clustering, applied at the second step. We choose to use the average distance to have the results represented in Figure 3 which are quite similar to what we obtained with our method.
% \begin{center} \begin{figure}[!thbt] \centering
%
%  		 \includegraphics[width=0.45\linewidth, height=4cm]{Bhclust10.eps}

%  		 \hfill
%  		 \includegraphics[width=0.45\linewidth, height=4cm]{Bhclust13.eps}
% 		 \hfill
% 		 \includegraphics[width=0.45\linewidth, height=4cm]{Mhclust10.eps}
% 		 \hfill
% 		 \includegraphics[width=0.45\linewidth, height=4cm]{Mhclust13.eps}
% 		 \hfill \parbox{.45\linewidth}{\centering\small a. $k=10$.} \hfill
% 		 \parbox{.45\linewidth}{\centering\small b. $k=13$.} \hfill
% 				 \caption{Mixed clustering results for iris data set.}
%  \label{1013} \end{figure} \end{center}
\section{Conclusion}
Ultimately, we have introduced a new clustering method using the hierarchical paradigm in order to implement uncertainty in the belief function framework. This method puts the emphasis on the fact that one object may belong to several clusters. It seeks to merge clusters based on its pignistic probability. Our method was proved on data sets and the corresponding results have clearly shown its efficiency. The algorithm complexity has revealed itself as the usual problem of the belief function theory. Our future work will be devoted to focus on this peculiar problem.
\end{document}